\documentclass[conference]{IEEEtran}
\usepackage{cite}
\ifCLASSINFOpdf
\usepackage[pdftex]{graphicx}
\else
\fi
\usepackage{amsmath}
\usepackage{array}
\usepackage{fixltx2e}
\usepackage{stfloats}
\usepackage{amsmath,amssymb}
\usepackage{url}
\usepackage{xcolor}
\newcommand{\bb}[1]{\textcolor{black}{#1}}

\begin{document}
\title{Proximal Policy Optimization \\ with Continuous Bounded Action Space \\ via the Beta Distribution}

\author{\IEEEauthorblockN{Irving G. B. Petrazzini}
\IEEEauthorblockA{Automation and Systems Engineering\\
Federal University of Santa Catarina\\
Florianópolis, Santa Catarina, 88040-900\\
Email: irving.petrazzini@posgrad.ufsc.br}
\and
\IEEEauthorblockN{Eric A. Antonelo}
\IEEEauthorblockA{Automation and Systems Engineering\\
Federal University of Santa Catarina\\
Florianópolis, Santa Catarina, 88040-900\\
Email: eric.antonelo@ufsc.br}
}

\maketitle

\begin{abstract}
Reinforcement learning methods for continuous control tasks have evolved in recent years generating a family of policy gradient methods that rely primarily on a Gaussian distribution for modeling a stochastic policy.
  However, the Gaussian distribution has an infinite support, whereas real world applications usually have a bounded action space. This dissonance causes an estimation bias that can be eliminated if
  the Beta distribution is used for the policy instead, as it presents a finite support.
  In this work, we investigate how this Beta policy performs when it is trained by the Proximal Policy Optimization (PPO) algorithm on two continuous control tasks from OpenAI gym. For both tasks, the Beta policy is superior to the Gaussian policy in terms of agent's final expected reward, also showing more stability and faster convergence of the training process. For the CarRacing environment with high-dimensional image input, the agent's success rate was improved by 63\% over the Gaussian policy.  
\end{abstract}


%
\IEEEpeerreviewmaketitle

\section{Introduction}

\bb{
Deep Reinforcement Learning (RL) has achieved unprecedented results on challenging high-dimensional continuous state-space problems, 
surpassing human performance in 29 out of 49 Atari 2600 games in \cite{mnih2015human}, for instance. 
Later, AlphaGO, an agent that combines reinforcement learning and Monte Carlo balanced search tree algorithms with self play was able beat Lee Sedol, a 9th-dan, world champion \cite{silver2017mastering}.
In this context, Convolutional Neural Networks (CNNs) \cite{lecun2015deep} serve as function approximators for the Q-value function, since they can efficiently process image inputs and learn useful feature representations from these high-dimensional continuous state-space domains.}

\bb{
Handling discrete action spaces in a deep reinforcement task usually resumes to defining an output layer of a neural network that has the same dimension of the action space. If this space is small, an action can be easily drawn from the distribution yielded by the layer's activation. Otherwise, finding the best action for high-dimensional or continuous action spaces constitutes an expensive optimization process per se, which needs to be run inside another loop, the agent-environment cycle.
}

\bb{
Many interesting real-world problems such as control of robotic arms and autonomous cars require a continuous action space.
Instead of modeling the state-action Q-value function, model-free continuous control via reinforcement learning is made possible by directly optimizing a policy function which maps states to probability distributions over continuous action spaces.
This family of policy gradient methods have undergone important advancements allowing for high-dimensional continuous state spaces (such as images) and continuous action spaces
\cite{schulman2015trust,schulman2017proximal,mnih2016asynchronous,wang2016sample}. 
}

\bb{
To model a stochastic policy, these methods choose the Gaussian distribution $\mathcal{N}(\mu,\,\sigma^{2})$, whose parameters $\mu$ and $\sigma^{2}$ are to be estimated as outputs of a deep neural network. However, many real-world applications have bounded action spaces, usually owing to physical constraints, e.g., by the joints of a humanoid robot or manipulator, and by the accelerator and steering direction of a vehicle.
Thus, in these cases, this Gaussian policy, which has infinite support, introduces a estimation bias since it can give a nonzero probability for \textit{actions} outside the valid action space.
}

\bb{
Recently, \cite{chou2017improving} proposes to model the stochastic policy with the Beta distribution, with parameters $\alpha$ and $\beta$, such that the resulting policy has a suitably bounded action space, that presents no bias as the previously considered Gaussian distribution. Instead of the mean and variance, now the outputs of the neural network represent the policy parameters $\alpha$ and $\beta$.
The Beta distribution can be used with any on- or off-policy methods such as
Trust Region Policy Optimization (TRPO) \cite{schulman2015trust} and Actor-Critic Experience Replay (ACER) \cite{wang2016sample}.
}

\bb{So far, the Beta distribution has been evaluated only for TRPO and ACER on a variety of problems. 
Proximal Policy Optimization (PPO),  which evolved from TRPO but has a much simpler implementation and a similar performance to ACER, still lacks experimentation with the Beta distribution. 
\bb{This is the first work to report experiments on PPO with the Beta distribution on RL applications with high-dimensional observation spaces}.
Besides, our investigation focus on two continuous control applications from OpenAI Gym, the Lunar Lander and the Car Racing, both of which were not considered in \cite{chou2017improving}.
}

\bb{The benefits of the approach are better stability and faster convergence of the training process.
Furthermore, because the estimation bias is absent, the final learned Beta policy is superior to the final Gaussian policy.
We also report results better than state-of-the-art on the Car Racing problem.
}

\section{Background}

\subsection{Markov decision process}
We model our continuous control reinforcement learning task as a finite Markov decision process (MDP). An MDP consists of a state space $S$, an action space $A$, an initial state ${s}_{0}$, and a 
\bb{reward function $r(s, a) : S \times A$ that emits a scalar value for any transition from state $s$ taking action $a$. }
At each time step $t$, the agent selects an action $a_{t+1}$ according to a policy \bb{$\pi$, i.e.,  $a_{t+1}=\pi(s_{t})$, 
such that agent's future expected reward is maximized. }
A stochastic policy can be described as a probability distribution of taking an action $a_{t+1}$ given a state $s_{t}$  denoted as $\pi(a|s) : S \to A.$ A deterministic policy can be obtained by taking the expected value of the policy $\pi(a|s)$.

\subsection{Policy Gradient Methods}
\bb{
Value-based reinforcement learning methods first learn to approximate a value function $Q(s,a)$. The policy is obtained by finding the action that maximizes the latter, e.g., $\pi(s)=\arg\max_a Q(s,a)$.
On the other hand, policy gradient methods optimize directly an parametrized policy $\pi_{\theta} (a|s)$ that can model Categorical or Continuous actions for discrete and continuous spaces, respectively. 
}

For a given scalar performance measure 
$\textit{L}(\theta) =  v_{\pi_{\theta}}(s_{0})$, \bb{where $v_{\pi_{\theta}}$ is the true value function for $\pi_{\theta}$, the policy determined by $\theta$}, performance is maximized through gradient ascent on $\textit{L}$

\begin{align}
L(\theta) = \int_{S} \rho^{\pi}(s) \int_{A} \pi_{\theta}(s,a) r(s,a) da ds
\label{eq:perf_measure}
\end{align}
\begin{align}
= \mathbb{E}_{s\sim\rho^{\pi},a\sim\pi_{\theta}}[r(s,a)]
\end{align}

\begin{align}
\theta_{t+1} = \theta_{t} + \alpha \widehat{\nabla_{\theta} \textit{L}(\theta_{t})}
\label{eq:param_update}
\end{align}
\bb{where 
$
\rho^{\pi}(s)=\sum_{t=0}^{\infty}\gamma^{t}p(s_{t}=s)
$
is the unnormalized discounted state visitation frequency in the limit \cite{sutton2000policy} and $\alpha$ is the learning rate.}

\subsection{Proximal Policy Optimization}

\bb{Proximal Policy Optimization \cite{schulman2017proximal} is one of the most commonly used policy gradient methods. Among the several variants for the performance measures available, we consider the clipped surrogate objective as in \cite{schulman2017proximal}, \bb{as follows}:
}
\begin{align}  
\textit{L}_{t}^{CLIP}(\theta)= \hat{\mathbb{E}}_{t}\left[\min(r_{t}(\theta) \hat{A}, \mathrm{clip}(r_{t}(\theta),1-\epsilon, 1+\epsilon) \hat{A_{t}}) \right ]
\end{align}
\bb{where $\theta_\mathrm{old}$ is the vector of policy parameters before the update;  $r_{t}(\theta)$ denotes the probability ratio $\frac{\pi_{\theta}(a_t|s_t)}{\pi_{\theta_\mathrm{old}(a_t|s_t)}}      $; 
$\epsilon$ is a hyperparameter used to clip the probability ratio by $\mathrm{clip}(r_{t}(\theta),1-\epsilon, 1+\epsilon)$, avoiding large policy updates \cite{schulman2017proximal}; 
and $\hat{A}_{t}$ is an estimator of the advantage function at timestep $t$, which weights the ratio $r_{t}(\theta)$.
Here, $\hat{\mathbb{E}}_{t}$ denotes an empirical average over a finite set of samples.
}

\bb{The implementation of policy gradient considers a} truncated version of the Generalized Advantage Estimator (GAE), as in \cite{schulman2015high}:

\begin{align}
\hat{A}_{t} = \delta_{t} + (\gamma\lambda)\delta_{t+1} + ... + ... + (\gamma\lambda)^{T-t+1}\delta_{T-1}
\end{align}
\begin{align}
\delta_{t} = r_{t} + \gamma V(s_{t+1}) - V(s_{t}),
\end{align}
where the policy is run for $T$ timesteps (with $T$ less than the episode size).
As commonly used in the literature, $\gamma$ and $\lambda$ are discount factor and GAE parameter, respectively. To perform a policy update, each of $N$ (parallel) actors collect $T$ timesteps of data. Then we construct the surrogate loss on these $NT$ timesteps of data, and optimize it with ADAM algorithm \cite{kingma2014adam} 
with a learning rate $\alpha$, in mini-batches of size $m \leq NT$ for $K$ epochs.
Notice that $V_\theta(s)$ in GAE is learned simultaneously in order to reduce the variance of the advantage-function estimator.

\bb{Once we use a neural network architecture that shares parameters between the policy and value function, we must use a loss function that combines the policy surrogate and a value function error term. This objective is further augmented by adding an entropy term to ensure sufficient exploration. Combining these terms, we obtain the following objective, which is (approximately) maximized at each iteration \cite{schulman2017proximal}:
}
\begin{align}
    \textit{L}_{t}^{CLIP+VF+S}(\theta)=\hat{\mathbb{E}}_t \left[
    \textit{L}_{t}^{CLIP}(\theta) -
    c_{1} \textit{L}^{VF}_{t}(\theta)+
    c_{2} \textit{S}[\pi_{\theta}](s_{t}) 
    \right ],
\end{align}
where $S$ denotes an entropy bonus; $L_{t}^{VF}$ is the value function (VF) squared-error loss 
$(V_{\theta}(s_{t}) -  V_t^\mathrm{targ} )^{2}$, with $V_t^\mathrm{targ} = r_{t} + \gamma V_{\theta}(s_{t+1})$; and $c1$, $c2$ are coefficients for the VF loss and entropy term, respectively.

\subsection{Gaussian Distribution}

The Gaussian distribution is defined by the \bb{following} probability density function:

\begin{align}
f(x, \mu, \sigma) = \frac{1}{\sigma\sqrt{2\pi}} 
 \exp\left( -\frac{1}{2}\left(\frac{x-\mu}{\sigma}\right)^{\!2}\,\right)
\end{align}
whose parameters $\mu$ and $\sigma$ are to be estimated by a deep neural network that models a so-called Gaussian policy, i.e., a parametrized policy $\pi_{\theta}(a|s) \sim \mathcal{N}(\mu,\,\sigma^{2})\,$.

Therefore, when acting in stochastic mode, the agent samples the policy whereas in deterministic mode, $\pi(a|s) = \mu$. 
Since the Gaussian distribution has an infinite support, \bb{these sampled} actions are clipped to the agent's bounded action space.

\subsection{Beta Distribution}

\begin{figure}[!h]
\centering
\includegraphics[width=0.5\textwidth]{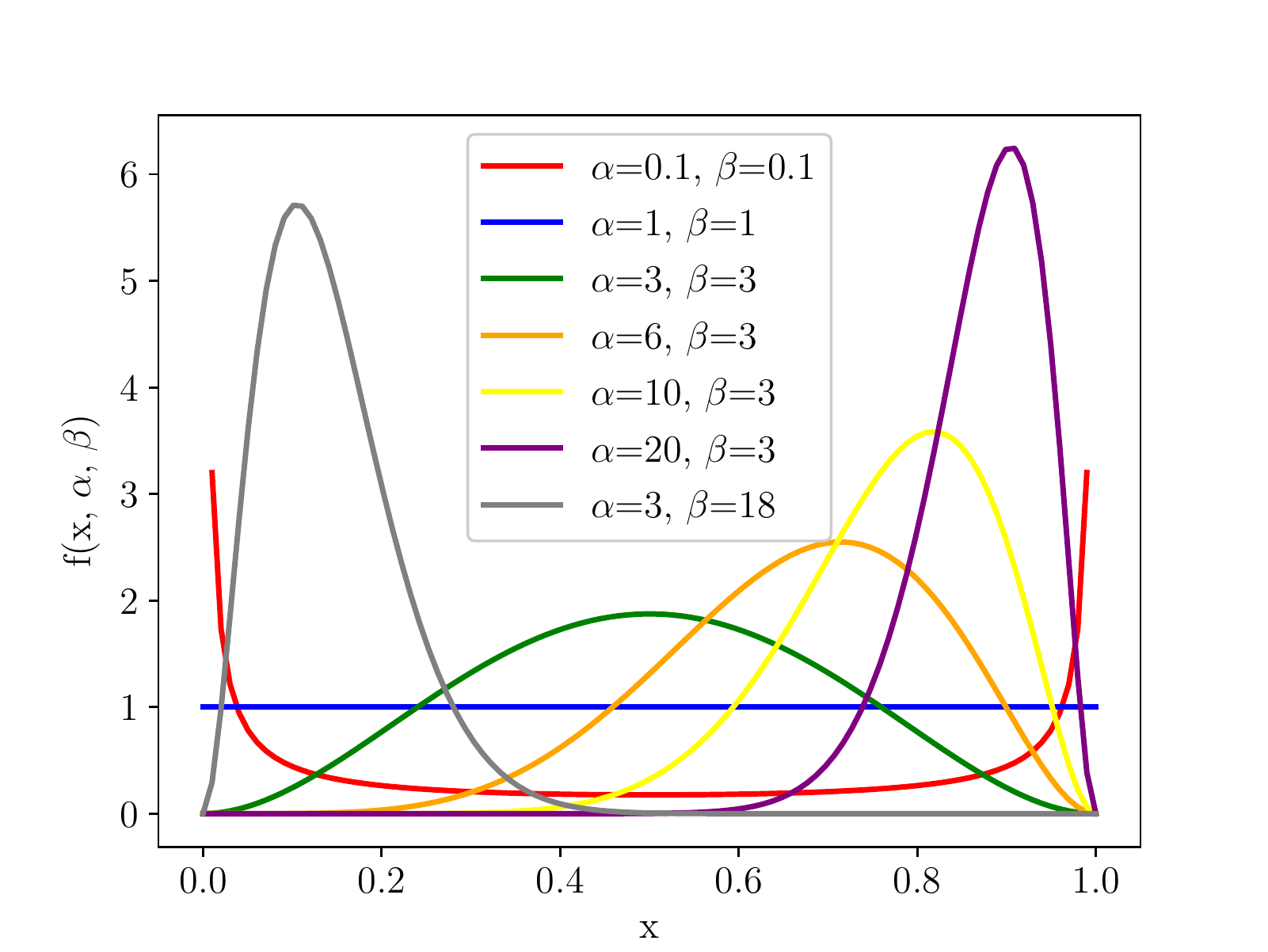}
\DeclareGraphicsExtensions.
\caption{Beta probability density function for different $\alpha,\beta$ pairs}
\label{fig:beta_plot}
\end{figure}

\bb{The Beta distribution \bb{has finite support} and can be intuitively understood as the probability of success, where $\alpha - 1$ and $\beta - 1$ can be thought of as the counts of successes and failures from the prior knowledge, respectively. For a random variable $x \in [0, 1]$, the Beta 
probability density function is given by:}
\begin{align}
 h(x:\alpha,\beta) = \frac{\Gamma(\alpha 
\beta)}{\Gamma(\alpha)\Gamma(\beta)}x^{\alpha - 1}(1 - x)^{\beta - 1}, 
\end{align}
\bb{where $\Gamma(.)$ is the Gamma function, which extends the factorial to real numbers. For $\alpha, \beta>1$, the distribution is uni-modal, as illustrated in Fig. \ref{fig:beta_plot}. When acting deterministically, \bb{the Beta policy outputs} $\pi_\theta(a|s) = \alpha/(\alpha + \beta)$. 
\bb{The $\alpha, \beta$ parameters that define the shape of the function are obtained as outputs of a deep neural network representing the parametrized policy $\pi_\theta(a|s)$.
}}

\subsection{Bias due to boundary effect}
\label{sec:bias}

Modeling a bounded action space by a probability distribution with infinite support \bb{possibly} introduces bias. As a result, the biased gradient imposes additional difficult in finding the optimal policy using reinforcement learning. 
The policy gradient estimator to optimize the parameters $\theta$ in (\ref{eq:param_update}), using $Q$ as the target, can be obtained by differentiating (1), as follows:
\begin{align}
\nabla_{\theta} \textit{L}(\theta_{t})=\int_{\mathbb{S}} \rho^\pi(s)
\int_{\mathbb{A}}\pi_{\theta}(a|s)\nabla_{\theta}\log\pi_{\theta}(a|s)Q^{\pi}(s|a)dads
\label{eq:L_grad_comp}
\end{align}
where  $Q^{\pi_{\theta}}(s,a)$ is a state-action value function for a policy $\pi_{\theta}$. Thus, the policy gradient estimator using $Q$ as the target is given by:

\begin{align}
g_{q}=\nabla_{\theta} \log \pi_{\theta}(a|s)Q^{\pi_{\theta}}(s,a)
\end{align}
This gradient is estimated by averaging $n$ samples with a fixed policy $\pi_{\theta}$, so that
\begin{align}
\nabla_{\theta} \textit{L}(\theta_{t}) = \frac{1}{n}\sum_{i=1}^{n}g_{q}  \rightarrow \mathbb{E}[g_{q}]=\nabla_{\theta} \textit{L}(\theta_{t}) \text{    as    } n \to \infty
\end{align} 

Let $A= [-h, h]$ be an uni-dimensional action space, with $a \in A $. 
In the case of an infinite support policy, an action $a' \notin A$ \bb{is eventually sampled}, to which the environments responds as if the action \bb{is either $h$ or $-h$}. 
\bb{The biased policy gradient estimator would be given by $g_{q}'=\nabla_{\theta}\log\pi_{\theta}(a|s)Q^{\pi}(s|a')$ in this case}. 
Besides, focusing on the inner integral of (\ref{eq:L_grad_comp}), the bias is computed as follows (also shown in \cite{chou2017improving}): 

$
\mathbb{E}[g_{q}'] - \nabla_{\theta}L(\theta)
$
\begin{align}
&= \mathbb{E}_s \left [   \int_{-\infty}^{\infty}\pi_{\theta}(a|s)\nabla_{\theta}\log \pi(a|s)Q^{\pi}(s,a')da\right ] - \nabla_{\theta}J(\theta) \notag \\
&= \mathbb{E}_s \left [\int_{-\infty}^{-h}\pi_{\theta}(a|s)\nabla_{\theta}\log \pi_{\theta}(a|s)[Q^{\pi}(s,-h)- Q^{\pi}(s,a)]da  \right. \notag \\
&+ \left. \int_{h}^{\infty}\pi_{\theta}(s|a)\nabla_{\theta}\log \pi_{\theta}(a|s)[Q^{\pi}(s,h)-Q^{\pi}(s,a)]da\right ]
\end{align}

These last two integrals evaluate to zero if the policy's distribution support is within the action space $A$.

\section{Experiments}

\bb{This section presents results for the proximal policy gradient method (PPO) on two continuous control problems from OpenAi gym \cite{1606.01540}: the LunarLanderContinuous-v2 with low-dimensional state space; and the CarRacing-v0 with high-dimensional image input (Table~\ref{tab:envs}). }


\bb{For all architectures, the last two layers output two 2-dimensional real vectors.  
For the Gaussian distribution, each dimension of the policy outputs its mean $\mu$ and its standard deviation $\sigma$, whereas for the Beta distribution the network outputs its parameters $\alpha$ and $\beta$ \textgreater 1.
\bb{Here, $1$} is added to a softplus layer $log(1+exp(x))$ to ensure both $\alpha$ and $\beta$ are larger than $1$. 
\bb{Our implementation for PPO was based on a popular reinforcement learning library found in \cite{pytorchrl}.} }

\begin{table}[h!]
    \centering
        \caption{Environments}    
        \label{tab:envs}
    \begin{tabular}{lcc}
    \hline
    Environment  &$ \left \| \mathcal{S} \right \| $&$ \left \| \mathcal{A} \right \| $\\
    \hline
    LunarLander  & 8       & 2 \\
    CarRacing    & 96x96x3 & 3 \\
    \hline
    \end{tabular}
\end{table}

For each environment, we trained five models using different seeds for both the Gaussian and Beta distributions for a fixed number of total time steps.  After completing the training, each model was evaluated in 100 consecutive episodes in both stochastic mode (sampling from the distribution) and deterministic mode (using the average of each distribution as the optimal action). 

\bb{
The hyperparameters for PPO
can be seen in Table~\ref{tab:hyperparams} for both control problems.
Notice that the PPO configuration for the Lunar Lander was adapted from the one used for the MuJoCo environment in \cite{schulman2017proximal},
whereas for the CarRacing, the parameters found in
Atari \cite{mnih2015human} were used as a starting point. 
}

\begin{table}[h!]
    \centering
        \caption{Hyperparameters for training }    
        \label{tab:hyperparams}
    \begin{tabular}{lcc}
    \hline
                                        &  Lunar Lander              &  CarRacing  \\ \hline
    Horizon (T)                         & $2048$                     & $500$                     \\
    Parallel environments ($N$)         & $1 $                       & $8$                       \\    
    Adam step size (lr)                 & $3 \times 10^{-4} \times \alpha$    & $2.5 \times 10^{-4} \times \alpha$   \\
    Number of PPO epochs (K)            & $10 $                      & $10$                      \\
    Mini-batch size (m)                 & $32  $                     & $64$                      \\
    Discount ($\gamma$)                 & $0.99$                     & $0.99$                    \\
    GAE parameter ($\lambda$)           & $0.95   $                  & $0.95 $                   \\
    Clipping parameter ($\epsilon$)     & $0.2  $                    & $0.1  $                   \\
    Value Function coefficient ($c_{1}$)& $0.5 $                     & $0.5  $                   \\
    Entropy coefficient ($c_{2}$)       & $0 $                       & $0.01  $                  \\
    Total timesteps                     & $10^6$                     & $5 \times 10^6$              \\
    \hline 
    \multicolumn{3}{c}{$\alpha$ is linearly annealed from 1 to 0 over the course of learning}
    \end{tabular}
    
\end{table}

\subsection{LunarLanderContinuous-v2}
\label{sec:exp:lunarlander}

\begin{figure}[b!]
\centering
\includegraphics[width=0.36\textwidth]{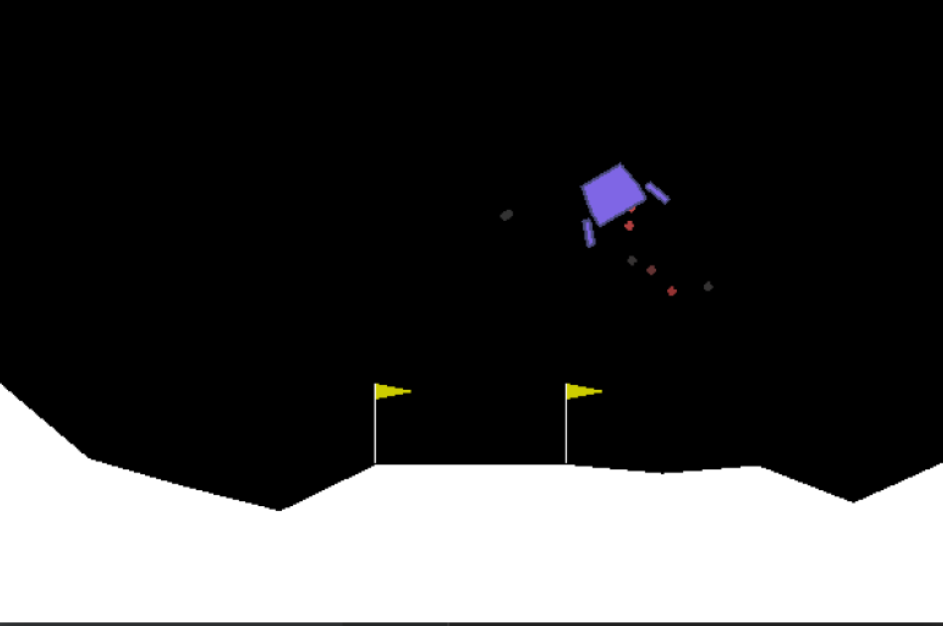}
\caption{LunarLanderContinuous-v2 Environment}
\label{fig:LLC_screen}
\end{figure}

\bb{The LunarLanderContinuous-v2 environment simulates the landing of a space module on the moon. The overall objective corresponds to landing the module on the lunar surface delimited by two flags, approaching zero speed at the final step (Fig.~\ref{fig:LLC_screen}). 
It has an unbounded, 8-dimensional observation space and a 2-dimensional action space. 
The actions are the main engine throttle and the secondary engine throttle, both bounded in the interval $[0, 1]$.  
The agent loses points for firing up the engines and for crashing (landing at high speed). The simulation is considered solved if the agent manages to score at least 200 points \cite{klimov2016LunarLander}. }

\bb{
The agent follows an actor-critic framework,
where the actor $\pi_\theta(a|s)$ consists of a neural network made of 3 fully-connected layers of 64 units each, with $\tanh$ activation functions. 
The output layer has 2 linear neurons to model either the Gaussian or the Beta distribution over the actions.
The critic $V_{\theta_v}(s)$ does not share layers with the actor, but has an equivalent architecture  of 3 hidden layers with only
one output neuron which represents the value function.
}

\subsection{CarRacing-v0}

The CarRacing-v0 environment \cite{klimov2016CarRarcing} simulates an autonomous driving environment in 2D. For each episode, a random track with 12 curves is generated. Each track is comprised of N tiles, with N ranging from 250 to 350. The agent receives 1000/N points for visiting each tile and loses 0.1 point for each frame. The episode ends in one of three situations: 

\begin{enumerate}
  \item Agent visited all tiles 
  \item Agent does not visit all tiles in 1000 frames
  \item Agent gets too far way from the track and falls in the abiss (-100 points added) 
\end{enumerate}

\begin{figure}[b!]
\centering
\includegraphics[width=0.32\textwidth]{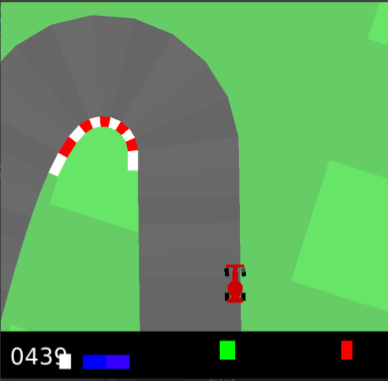}
\caption{CarRacing-v0 Environment}
\label{fig:CR_screen}
\label{fig:carracing_env}
\end{figure}

Therefore, if the agent visits all tiles in 732 frames, the reward is 1000 - 0.1*732 = 926.8 points. Should the agent miss one or more tiles in its first lap attempt, the episode keeps on until the agent visit missing frame or the time limit is reached. 
\bb{The task is considered to be solved if the agent is able to get an average reward of at least 900 points in 100 consecutive trials (episodes).}

\bb{The observation space consists of top down images (Fig.~\ref{fig:carracing_env}) of 96x96 pixels and three (RGB) color channels. The latest four image frames were stacked and given as input to the agent's network after rescaling and preprocessing them to gray scale (totalling 84x84x4 input dimensions).}
The action space has three dimensions: one encodes the steering angle and is bounded in the interval $[-1, +1]$. The other two dimensions encode throttle and brake, both bounded to $[0, 1]$.

For our implementation, throttle and brake have been merged on a single dimension so that on a given step, the agent does not simultaneously accelerates and brakes.  
We believe this is a more representative structure of real world systems: separated control inputs (throttle/brake) but single activation mechanism (right foot). 
\bb{In practice, one output neuron is responsible for both actions, making the output of the agent to be a two-dimensional vector.
}
With this approach, we were able to make the agent learn effectively, 
mainly because it does not enter a deadlock state resulting from accelerating and braking at the same time. If we did not follow this approach, learning to control the vehicle would not take place. 
\bb{So far, we were not able to find other work in the literature that takes advantage on the aforementioned approach.} Also, notice that we have not changed the original reward signal as some other works might have done.

\bb{The actor-critic network resembles that of \cite{mnih2015human}
with respect to the shared encoder base comprised of the first 3 convolutional layers.
Instead of connecting directly to the output layer as in \cite{mnih2015human}, the shared base has an additional fully connected (FC) layer with 512 units.
The critic $V_{\theta_v}(s)$  specializes further with its exclusive 1 FC layer of 512 units, that connects to a final output. 
The actor $\pi_\theta(a|s)$  has its own 2 FC layers with 512 units each on top of the shared base.
The output layer is equivalent to the one from Section ~\ref{sec:exp:lunarlander}, but its two neurons now refer to the steering angle or acceleration (brake/throttle). 
}
\\

    

\section{Results and Discussion}

\subsection{LunarLanderContinuous-v2}
For the LunarLanderContinuous-v2 environment, we observe that using a Beta distribution allow for both a faster convergence and higher total reward during training. 
Five agents were \bb{trained} with the same \bb{hyperparameters} and different seeds. 
\begin{figure}[b!]
\centering
\includegraphics[width=0.5\textwidth]{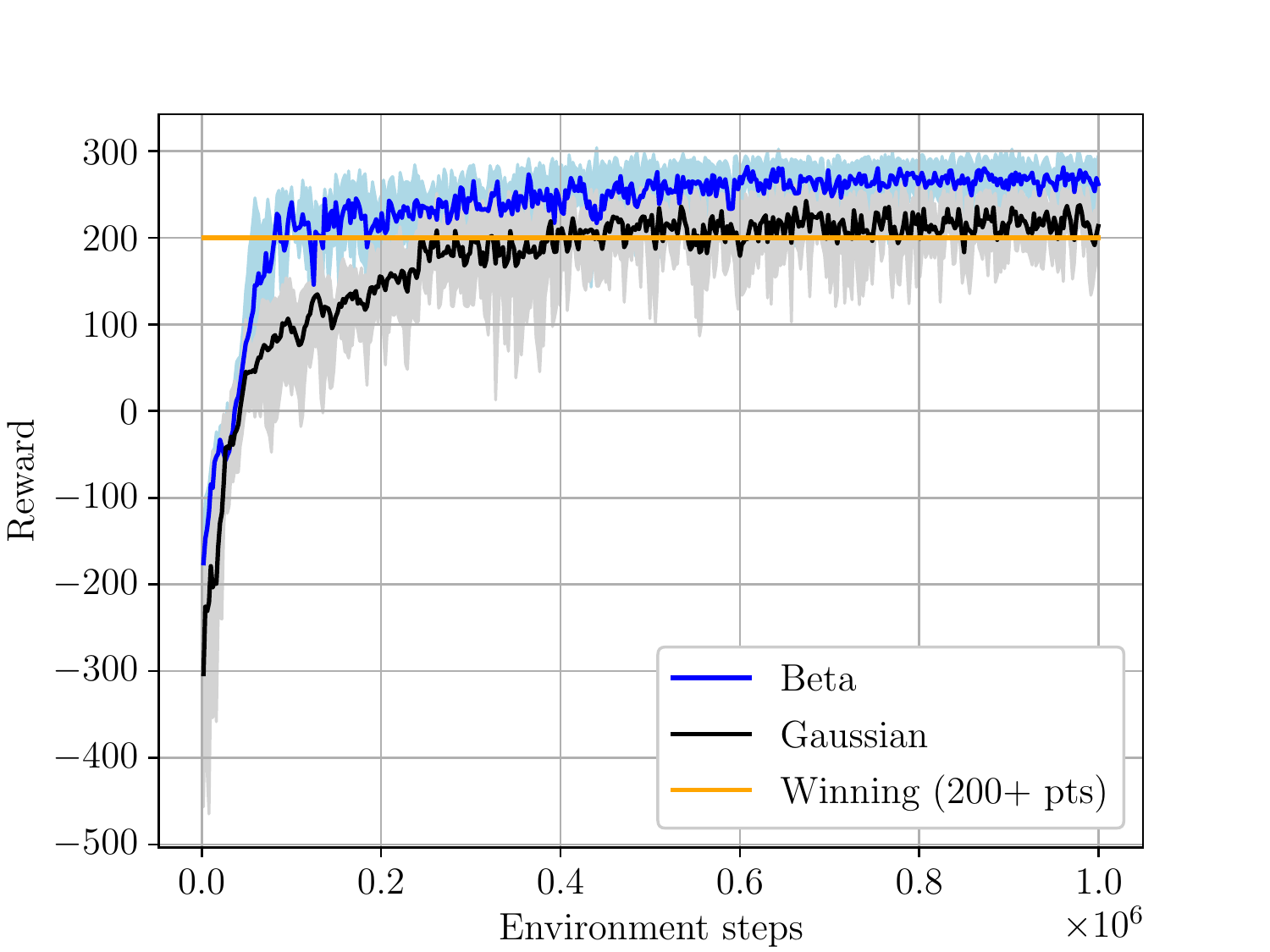}
\caption{\bb{Average rewards for five agents on the Lunar Lander task trained with Beta or Gaussian distribution for 1 million steps.
The solid line represents the mean
over a moving window of the previous 10 episodes for these five agents.
The shaded area represents the interval between the  minimum reward and maximum reward.
}
}
\label{fig:LLC_train}
\end{figure}

\begin{figure}[]
\centering
\includegraphics[width=0.45\textwidth]{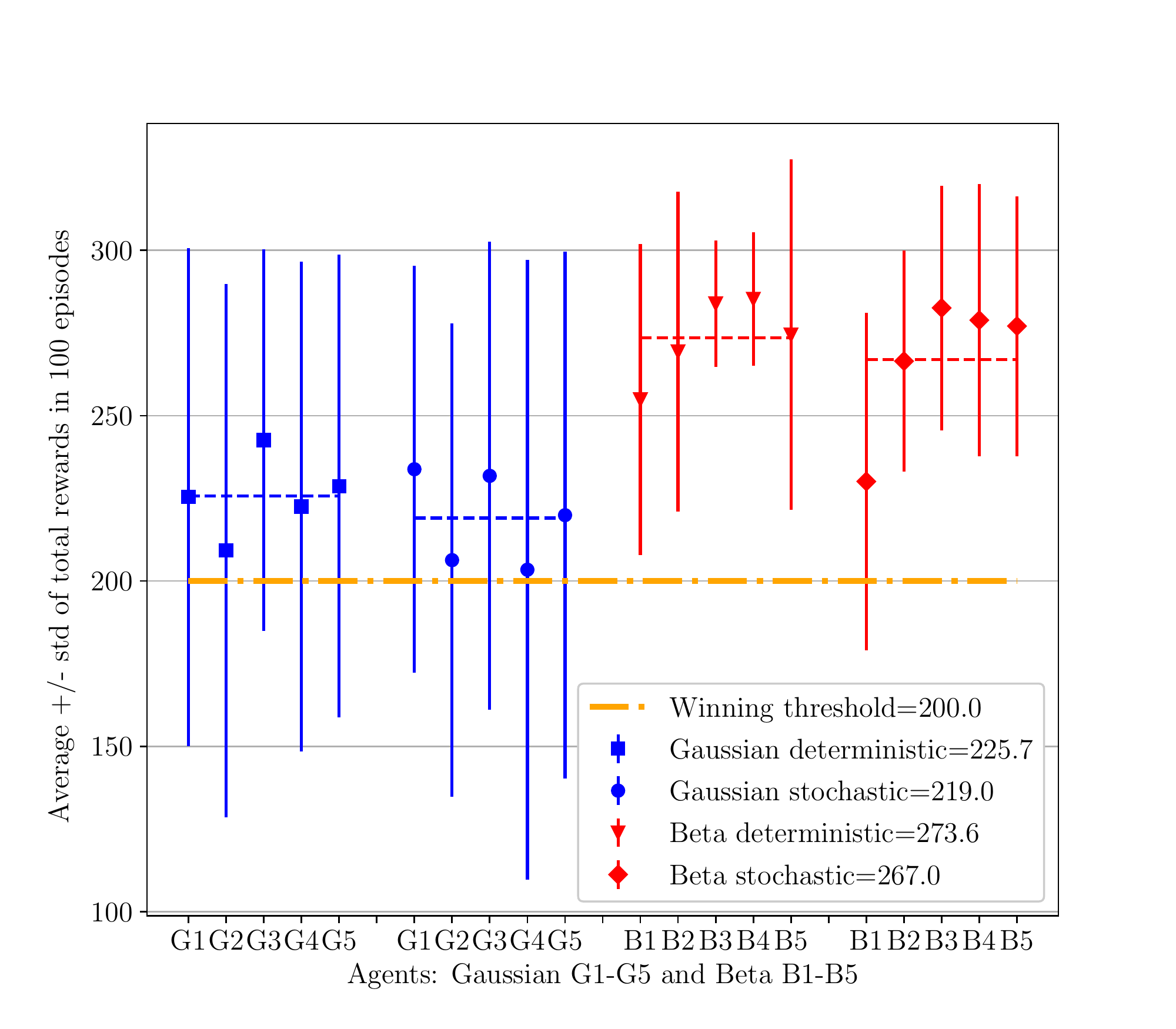}
\caption{Lunar Lander results: \bb{comparison of the Gaussian policy to the Beta policy in terms of the average rewards obtained by agents for 100 consecutive episodes after training.
In blue (red), 
the average reward and standard deviation for each one of 5 agents using
the Gaussian (Beta) policy.
Both deterministic and stochastic policies were employed for evaluation.
The winning threshold given by the horizontal black line represents the minimum threshold for successful completion of the task.
Agents powered by the Beta distribution achieved superior performance and less variance. 
}
}
\label{fig:LLC_evaluation}
\end{figure}

After a million times steps, training is frozen and we evaluated each agent for 100 episodes in deterministic mode (using the mean of the \bb{policy's distribution as} 
the action) and in stochastic mode (sampling the \bb{policy's distribution}). 

For the Gaussian distribution, we observe that the performance of the agents hovers around \bb{225.7 and 219.0} points for the deterministic and stochastic policies, respectively. 
For the Beta distribution, we observe the agents \bb{perform at 267.0 (deterministic) and  273.6 points (stochastic)}. 
It is worth noting that Agents B4 and B5, which were trained with the Beta distribution, were able to score \bb{at least} 200 points for all 100 episodes (Fig. \ref{fig:LLC_evaluation}) whereas the best agent trained with the Gaussian distribution (G3, deterministic policy) was able to score above the \bb{200 points} threshold for 92\% of the \bb{100} evaluation episodes.
We can also observe that the variance of the Gaussian policy is higher than that of the Beta policy, even at the latest training iterations (Fig.~\ref{fig:LLC_train}) or after training ends (Fig.~\ref{fig:LLC_evaluation}).

\subsection{CarRacing-v0}

\begin{figure*}[h!]
\centering
\includegraphics[width=0.75\textwidth]{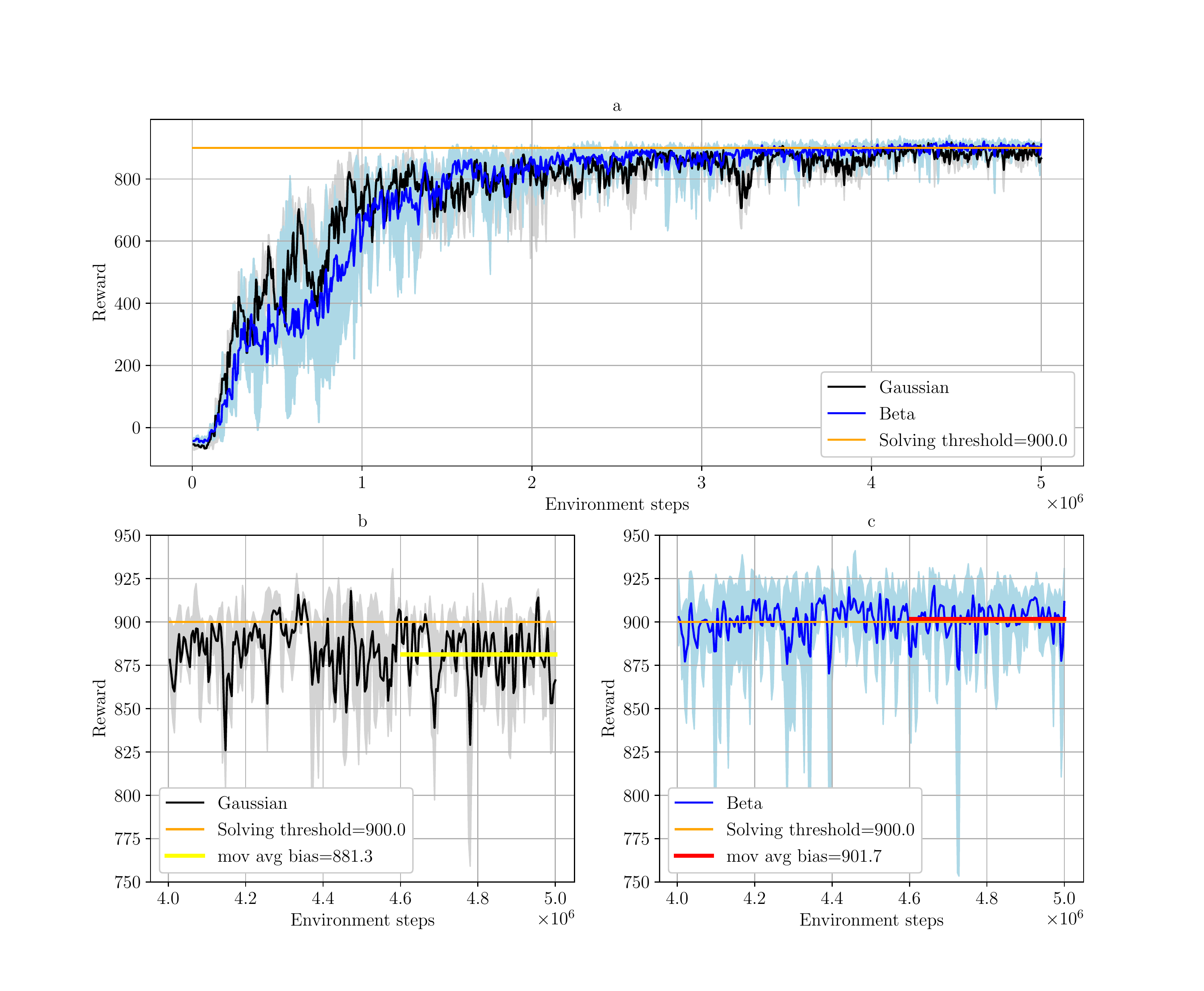}
\caption{
Average rewards for 5 agents with different seeds for the CarRacing environment, plotted equivalently to Fig.~\ref{fig:LLC_train}.
The final performance is shown on the bottom plots at a bigger scale,
for agents with Gaussian policy (bottom left) and Beta policy (bottom right).   
}
\label{fig:CarRacing_training}
\end{figure*}

For the CarRacing-v0 environment,
the number of agent-environment interactions was fixed to 5 million steps during training.
Afterwards, an evaluation of the agent's
performance takes place,
measured as the
average reward in 100 consecutive episodes. The task is
solved if this value is at least 900 points. 
We have observed that agents trained with Beta and Gaussian distributions have a similar convergence rate during training time. In Figure \ref{fig:CarRacing_training}, we show the average reward  over a moving window of 10 episodes, \bb{along the training process.} 
Each policy optimization takes in 500 environment steps across 8 parallel environments. 

\begin{figure}[h!]
\centering
\includegraphics[width=0.45\textwidth]{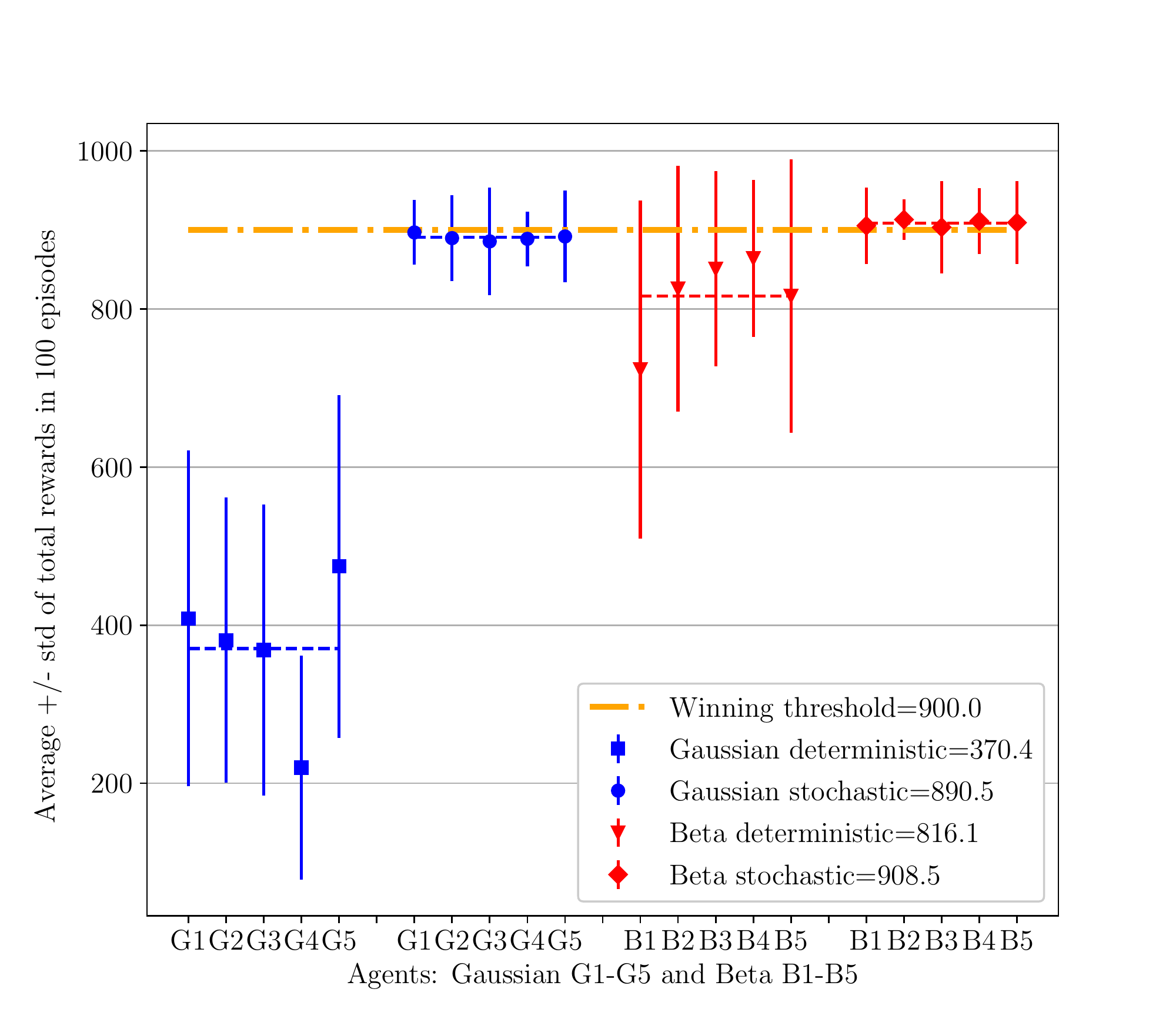}
\caption{Car Racing results after training: The evaluation followed the same procedure used for Lunar Lander (plots can be understood as in Fig.~\ref{fig:LLC_evaluation}). For this task, the stochastic policy clearly yielded better performance than the deterministic one. 
}
\label{fig:CR_evaluation}
\end{figure}
\begin{figure*}[]
\centering
\includegraphics[width=0.94\textwidth]{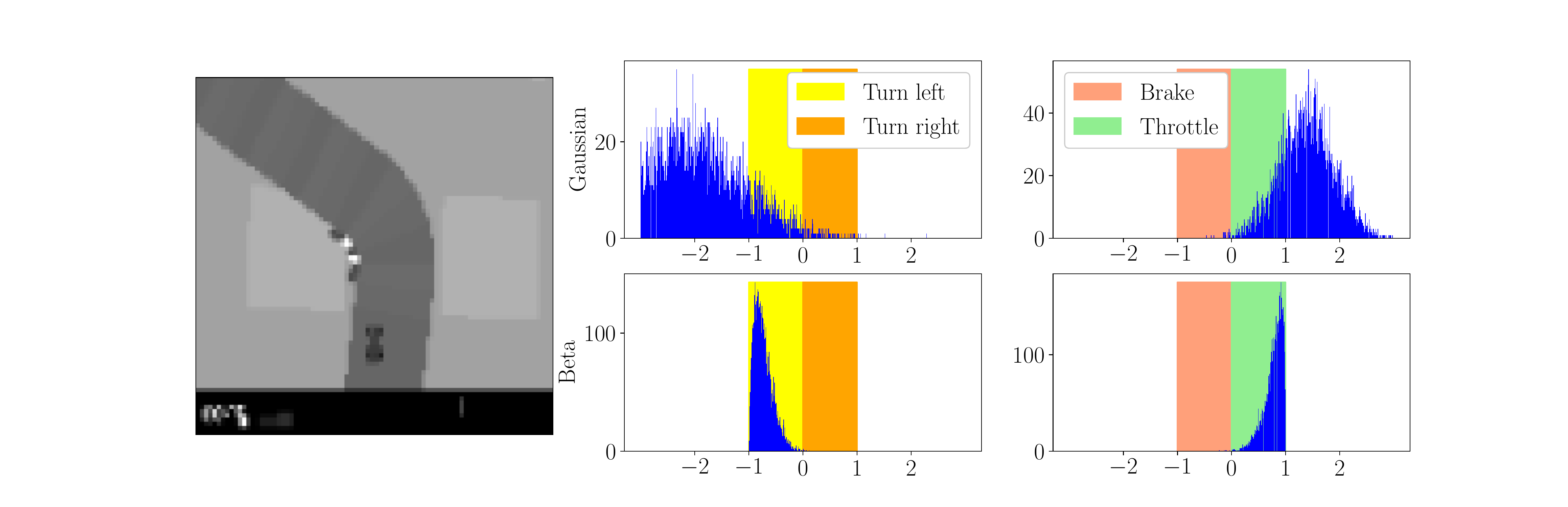}
\DeclareGraphicsExtensions.
\caption{Illustration of the Gaussian and Beta stochastic policy distributions in relation to the action space of the CarRacing environment. 
For a fixed observation $s$ (preprocessed image in left plot), we sampled the Gaussian and Beta policies for 5000 actions. 
For the Gaussian distribution, a significant portion of the actions fall out of the valid direction and brake/throttle range (both $[-1, 1]$), whereas for the Beta distributions, all actions fall within boundaries.}
\label{fig:policy_sampling}
\end{figure*}

\bb{Using the performance measure for 100 consecutive episodes training,
in Fig.~\ref{fig:CR_evaluation}, we show that the  
stochastic policy presented better average performance than the deterministic policy for both distributions.
For the Gaussian distribution, we observe that the five agents with the deterministic policy fail to follow the track, presenting an average score of 370.4 points that is much lower than the required 900 score points to solve the task. 
In stochastic mode, the policy presents an improved performance with an average score of 890.5 points, although in 38\% of the 100 episodes the agents were not able to pass the winning threshold.
%
For the Beta distribution, the agents' performance with the deterministic policy improves over the Gaussian policy by 320\%, with average score of 816.1 points.
These agents surpass the winning threshold in 26\% of the evaluation episodes. 
In the stochastic mode, all agents were able to score above the winning threshold in at least 60\% of the 100 of episodes played by each agent. 
All five agents with the Beta policy were able to successfully solve the game since each one of them reached a performance higher 900 points. This was not the case for the stochastic Gaussian policy, where each agent performed less than the threshold of 900 points.
The best performing agent, B2, consistently reached scores above the other five agents, and it's the chosen agent to compare our approach with other works in the literature in the next section.}
\bb{Fig. \ref{fig:policy_sampling} shows the resulting Gaussian and Beta policies at a specific timestep of the simulation, after training, when the car was about to turn left as it can be seen on the image fed to the policy network. The sampled distributions for both policies show that the Gaussian distribution, with its infinite support, falls outside the bounded action space, what is associated with the bias calculated in Section~\ref{sec:bias}. On the other hand, the Beta distribution fits well within the bounded action space, yielding an unbiased policy gradient estimator.}

\subsection{Considerations on the CarRacing-v0 environment and other approaches}

Simulation environments designed as test beds for reinforcement learning 
\bb{algorithms}
are primarily used in two ways: 

\begin{enumerate}
  \item To benchmark new algorithms or techniques without focusing particularly on a specific task;
  \item 
  To develop methods to solve a specific simulation task or benchmark, such as scoring more than 900 points on average in 100 consecutive runs for the CarRacing-v0 env, in an attempt to beat the other reported results.\\
\end{enumerate}

Although our primary objective was the former, we emphasize that our work happens to 
\bb{fulfill}
to the latter as well. OpenAI CarRacing-v0 Leaderboard \cite{openAI2016Leaderboard} hosts a series of self-reported scores. We compare our results only to those found in peer-reviewed articles (Table \ref{leaderboard}), 
\bb{since they}
provide a basis for comparison and discussion.

\begin{table}[b!]
    \centering
        \caption{CarRacing-v0 Leaderboard }    
        \label{tab:Leaderboard}
    \begin{tabular}{lcc}
    \hline
    Method  &  Average Evaluation Score  \\
    \hline
    \textbf{PPO with Beta (Ours)}                       & \textbf{913 +/- 26}  \\
    World models \cite{ha2018recurrent}                 & 906 +/- 21  \\
    Adapted DQN \cite{rodrigues2020optimizing}          & 905 +/- 24  \\
    Genetic Algorithms \cite{risi2019deep}              & 903 +/- 72  \\
    \hline
    PPO with Gaussian (Ours)                            & 897 +/- 41  \\
    Weight Agnostic NN \cite{gaier2019weight}           & 893 +/- 74  \\
    PPO \cite{jena2020augmenting}                       & 740 +/- 86  \\
    Random agent                                        & -32 +/- 6   \\
    \hline
    \label{leaderboard}
    \end{tabular}
\end{table}

Among the works that use Car Racing as a test bed, \cite{ha2018recurrent} claim to have been the first to solve the problem, using a recurrent world model. Other attempts included Deep Q-Networks with action-space discretization \cite{rodrigues2020optimizing} and Genetic algorithms \cite{risi2019deep}. 
\bb{Other work that uses the Car Racing environment for benchmarking other algorithms 
are \cite{gaier2019weight} and \cite{jena2020augmenting}, and have been included for reference. 
}

\section{Conclusions}

\bb{In this study, we observed that agents trained with PPO using a Beta distribution for the stochastic policy presented faster and more stable convergence of the training process (mainly for the Lunar Lander task), while their final performance was significantly superior to those trained with a Gaussian distribution.
Thus, the Beta distribution is better able to satisfy the requirements of real-world applications with bounded action spaces, overcoming the estimation bias of the Gaussian policy.}

\bb{Our results also show that continuous control with bounded action space for challenging car racing with random tracks and
a high-dimensionality of the observation space (based on images) is much facilitated when the Beta distribution is employed. In fact, the agent's success in this task is considerably affected by this approach, achieving the best score so far on the CarRacing-v0 Leaderboard among the published work in literature.
Finally, the results suggest that the Beta distribution should be a standard choice for those type of tasks.}

Originally proposed in \cite{chou2017improving}, the Beta distribution was tested in their work with TRPO/ACER on Atari games, which have high-dimensional observation space, but a discrete action space; and on robotic control tasks with a continuous action space and a low-dimensional observation space. In this work, we proposed to use the Beta distribution with PPO on high-dimensional image inputs and continuous action spaces.

\bb{We plan to extend these experiments to other types of reinforcement learning algorithms that are more sample efficient, in an attempt to verify if the Beta distribution transfers to other setups. Besides, experiments with more complex autonomous navigation in urban scenarios could benefit from the faster and more stable convergence as the training of end-to-end models is not a trivial task.}

\section*{Acknowledgment}

The authors would like to thank CAPES/Brazil for the financial sponsorship.



%

\bibliographystyle{IEEEtran}
\bibliography{ssci_ppo_beta.bbl}
\end{document}